\documentclass[10pt,twocolumn,letterpaper]{article}

\usepackage{cvpr}              % To produce the CAMERA-READY version
\usepackage{graphicx}
\usepackage{amsmath}
\usepackage{amssymb}
\usepackage{booktabs}
\usepackage{footmisc}  % Add this in the preamble
\usepackage[pagebackref,breaklinks,colorlinks]{hyperref}
\usepackage[normalem]{ulem}  % Enables underlining
\hypersetup{
    colorlinks=true,
    urlcolor=blue
}

\usepackage{multicol} 

\usepackage[capitalize]{cleveref}
\crefname{section}{Sec.}{Secs.}
\Crefname{section}{Section}{Sections}
\Crefname{table}{Table}{Tables}
\crefname{table}{Tab.}{Tabs.}
\newcommand{\mysection}[1]{\vspace{2pt}\noindent\textbf{#1}}

\def\cvprPaperID{*****} % *** Enter the CVPR Paper ID here
\def\confName{CVPR}
\def\confYear{2022}

\global\long\def\comma{\,,}%
\global\long\def\point{\,.}%

\begin{document}

\title{SoccerNet 2024 Challenges Results}
\author{
 \parbox{\linewidth}{\centering
 Anthony Cioppa$^{\dag1,2*}$, 
 Silvio Giancola$^{\dag2*}$, 
 Vladimir Somers$^{\dag3,4,5}$, 
 Victor Joos$^{\dag4}$, 
 Floriane Magera$^{\dag1,6}$,
 \\
Jan Held$^{\dag1}$, 
Seyed Abolfazl Ghasemzadeh$^{\dag4}$, 
Xin Zhou$^{\dag7}$, 
Karolina Seweryn$^{\dag8}$, 
Mateusz Kowalczyk$^{\dag8}$, 
Zuzanna Mróz$^{\dag8}$, 
Szymon Łukasik$^{\dag8, 32}$, 
Michał Hałoń$^{\dag8}$, 
Hassan Mkhallati$^{\dag9}$, 
Adrien Deliège$^{\dag1}$, 
 \\
Carlos Hinojosa$^{\dag2}$, 
Karen Sanchez$^{\dag2}$, 
Amir M. Mansourian$^{\dag10}$, 
Pierre Miralles$^{\dag11}$, 
Olivier Barnich$^{\dag6}$, 
Christophe De Vleeschouwer$^{\dag4}$, 
Alexandre Alahi$^{\dag5}$, 
Bernard Ghanem$^{\dag2}$, 
Marc Van Droogenbroeck$^{\dag1}$, 
Adam Gorski$^{26}$, 
Albert Clapés$^{19,20}$, 
Andrei Boiarov$^{18}$, 
Anton Afanasiev$^{18}$, 
Artur Xarles$^{19,20}$, 
 \\
Atom Scott$^{30}$, 
ByoungKwon Lim$^{15}$, 
Calvin Yeung$^{30}$, 
Cristian Gonzalez$^{16}$, 
Dominic Rüfenacht$^{12}$, 
 \\
Enzo Pacilio$^{16}$, 
Fabian Deuser$^{24}$, 
Faisal Sami Altawijri$^{14}$, 
Francisco Cachón$^{16}$, 
HanKyul Kim$^{15}$, 
 \\
Haobo Wang$^{31}$, 
Hyeonmin Choe$^{22}$, 
Hyunwoo J, Kim$^{25}$, 
Il-Min Kim$^{22}$, 
Jae-Mo Kang$^{22}$, 
 % \\
Jamshid Tursunboev$^{22}$, 
Jian Yang$^{17}$, 
Jihwan Hong$^{25}$, 
Jimin Lee$^{25}$, 
Jing Zhang$^{23}$, 
Junseok Lee$^{25}$, 
 % \\
Kexin Zhang$^{23}$, 
Konrad Habel$^{24}$, 
Licheng Jiao$^{23}$, 
Linyi Li$^{31}$, 
Marc Gutiérrez-Pérez$^{13}$, 
Marcelo Ortega$^{16}$, 
Menglong Li$^{29}$, 
Milosz Lopatto$^{26}$, 
Nikita Kasatkin$^{18}$, 
Nikolay Nemtsev$^{18}$, 
Norbert Oswald$^{24}$, 
 % \\
Oleg Udin$^{18}$, 
Pavel Kononov$^{18}$, 
Pei Geng$^{17}$, 
Saad Ghazai Alotaibi$^{14}$, 
Sehyung Kim$^{25}$, 
Sergei Ulasen$^{18}$, 
Sergio Escalera$^{19,20,21}$, 
Shanshan Zhang$^{17}$, 
Shuyuan Yang$^{23}$, 
Sunghwan Moon$^{22}$, 
 % \\
Thomas B. Moeslund$^{21}$, 
Vasyl Shandyba$^{18}$, 
Vladimir Golovkin$^{18}$, 
Wei Dai$^{29}$, 
WonTaek Chung$^{15}$, 
 % \\
Xinyu Liu$^{23}$, 
Yongqiang Zhu$^{29}$, 
Youngseo Kim$^{25}$, 
Yuan Li$^{28}$, 
Yuting Yang$^{23}$, 
Yuxuan Xiao$^{17}$, 
 % \\
Zehua Cheng$^{27}$, 
Zhihao Li$^{17}$
\\
\vspace{3mm}
$^{1}$~University of Liege (ULiège), 
$^{2}$~King Abdullah University of Science and Technology, 
$^{3}$~Sportradar, 
$^{4}$~UCLouvain, 
$^{5}$~EPFL, 
$^{6}$~EVS Broadcast Equipment, 
$^{7}$~Baidu Research, 
$^{8}$~NASK National Research Institute, 
$^{9}$~Université Libre de Bruxelles (ULB), $^{10}$~Sharif University of Technology, 
$^{11}$~Footovision, 
$^{12}$~Mobius Labs, 
$^{13}$~Institut de Robòtica i Informàtica Industrial, CSIC-UPC, 
$^{14}$~TAHAKOM, 
$^{15}$~AIBrain, 
$^{16}$Eidos.ai, 
$^{17}$~School of Computer Science and Engineering, Nanjing University of Science and Technology, 
$^{18}$~Constructor.tech, 
$^{19}$~Universitat de Barcelona, 
$^{20}$~Computer Vision Center, 
$^{21}$~Aalborg University, 
$^{22}$~Kyungpook National University, 
$^{23}$~Intelligent Perception and Image Understanding Lab, Xidian University, 
$^{24}$~University of the Bundeswehr Munich - Institute for Distributed Intelligent Systems (VIS), 
$^{25}$~Korea University, 
$^{26}$~Warsaw University of Technology, 
$^{27}$~University of Oxford, 
$^{28}$~China Science IntelliCloud Technology Co., Ltd., 
$^{29}$~Robo Space, 
$^{30}$~Graduate School of Informatics, Nagoya University, 
$^{31}$~Tongji University
$^{32}$~AGH University of Krakow}
}

\maketitle

\begin{figure*}
    \centering
    \includegraphics[width=\linewidth]{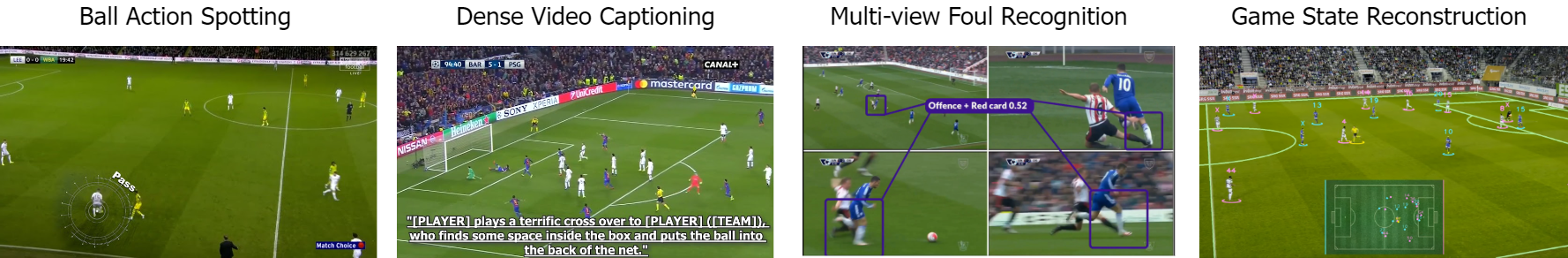}
    \caption{
In 2024, the challenges encompass four vision-based tasks.
(1) \textit{Ball Action Spotting}, focusing on precisely localizing when and which soccer actions related to the ball occur, 
(2) \textit{Dense Video Captioning}, focusing on describing the broadcast with natural language and anchored timestamps, 
(3) \textit{Multi-View Foul Recognition}, a novel task focusing on analyzing multiple viewpoints of a potential foul incident to classify whether a foul occurred and assess its severity, 
(4) \textit{Game State Reconstruction}, another novel task focusing on reconstructing the game state from broadcast videos onto a 2D top-view map of the field.
}
    \label{fig:graphical_abstract}
\end{figure*}
%%%%%%%%% ABSTRACT
\begin{abstract}

\renewcommand{\thefootnote}{} % Remove numbering in the footnote
\footnotetext{$^*$Denotes equal contributions and $^\dag$ denotes challenges organizers.}
\renewcommand{\thefootnote}{\arabic{footnote}}  % Restore numbering after the footnote

The SoccerNet 2024 challenges represent the fourth annual video understanding challenges organized by the SoccerNet team. 
These challenges aim to advance research across multiple themes in football, including broadcast video understanding, field understanding, and player understanding.
This year, the challenges encompass four vision-based tasks.
(1) \textit{Ball Action Spotting}, focusing on precisely localizing when and which soccer actions related to the ball occur, 
(2) \textit{Dense Video Captioning}, focusing on describing the broadcast with natural language and anchored timestamps, 
(3) \textit{Multi-View Foul Recognition}, a novel task focusing on analyzing multiple viewpoints of a potential foul incident to classify whether a foul occurred and assess its severity, 
(4) \textit{Game State Reconstruction}, another novel task focusing on reconstructing the game state from broadcast videos onto a 2D top-view map of the field.
Detailed information about the tasks, challenges, and leaderboards can be found at 
% \hyperref[https://www.soccer-net.org]{https://www.soccer-net.org}
\href{https://www.soccer-net.org}{\uline{https://www.soccer-net.org}}, with baselines and development kits available at 
% \hyperref[https://github.com/SoccerNet]{https://github.com/SoccerNet}. 
% {\hyperlink{https://github.com/SoccerNet}{\underline{https://github.com/SoccerNet}}}.
% \url{https://github.com/SoccerNet}.
% \href{https://github.com/SoccerNet}{https://github.com/SoccerNet}.
\href{https://github.com/SoccerNet}{\uline{https://github.com/SoccerNet}}.

\end{abstract}

\section{Introduction}
\label{sec:intro}

Sports video understanding has become a major field of computer vision research. To advance video analysis in sports, the SoccerNet dataset has introduced various tasks related to video understanding. In 2024, we featured four of these tasks in our annual open challenge competition to engage the academic and industrial research communities in achieving state-of-the-art performances. This paper details the outcomes of the SoccerNet 2024 challenges, highlighting the innovative solutions proposed by the participants.

\subsection{SoccerNet dataset history}
Giancola \etal introduced \textit{SoccerNet}~\cite{Giancola2018SoccerNet} in 2018 with the objective of publicly sharing a large-scale dataset for reproducible and open-source research and benchmark in soccer video understanding. Originally, the dataset contained 500 videos of complete broadcast soccer games, totaling nearly 800 hours of footage from the six major European championships (Serie A, La Liga, Premier League, Ligue 1, Bundesliga, and Champions League) spanning from 2014 to 2017.
With the dataset, Giancola \etal introduced the novel task of action spotting, focusing on the temporal localization of three major soccer actions: goals, cards, and substitutions.
Over the following years, the \textit{SoccerNet} dataset grew continuously, including more novel tasks and data.
First, Deliège~\etal introduced \textit{SoccerNet-v2}~\cite{Giancola2018SoccerNet}, which significantly increased the number of annotations for the action spotting task by including more classes such as penalties, clearances, and ball out of play, totaling $110{,}458$ actions across $17$ classes. Additionally, \textit{SoccerNet-v2} incorporated annotations for all camera changes, covering $13$ camera classes and $3$ transition classes. Each camera shot was further annotated to indicate if it contained a replay of an action, linking it to its corresponding action timestamp during the live feed. Alongside the enhanced dataset, three tasks were proposed: action spotting, camera shot segmentation and boundary detection, and replay grounding. The first version of the SoccerNet challenges was organized in 2021, focusing on action spotting and replay grounding.
In 2022, Cioppa \etal introduced \textit{SoccerNet-v3}~\cite{Cioppa2022Scaling} and \textit{SoccerNet-Tracking}~\cite{Cioppa2022SoccerNetTracking}. \textit{SoccerNet-v3} includes new multi-view image-based spatial annotations covering players, the ball, field lines, and goal parts, along with three new tasks: pitch localization, camera calibration, and player re-identification. \textit{SoccerNet-Tracking} introduced the task of multiple object tracking across single-view video clips, promoting long-term tracking and incorporating metadata such as jersey numbers and team affiliations.
In 2023, \textit{SoccerNet} received two enhancements. The first, \textit{SoccerNet-Captions}~\cite{Mkhallati2023SoccerNetCaption}, introduced natural language descriptions of events in broadcast games, establishing a new task for dense video captioning. The second, \textit{SoccerNet-MVFouls}~\cite{Held2023VARS}, presented a multi-view video dataset designed to recognise and characterise fouls.
In 2024, \textit{SoccerNet} introduced \textit{SoccerNet-Depth}~\cite{Leduc2024SoccerNetDepth}, the largest publicly available dataset for monocular depth estimation on team sports videos, \textit{SoccerNet-XFoul}~\cite{Held2024XVARS}, a novel multi-modal dataset containing more than $22$k video-question-answer triplets about refereeing decisions, and \textit{SoccerNet-GSR}~\cite{Somers2024SoccerNetGameState}, the first open-source sports
video dataset for game state reconstruction.

\subsection{SoccerNet 2024 challenges overview}

The 2024 edition of the SoccerNet challenges proposed four vision-based tasks: 
(1) \textit{Ball Action Spotting}, focusing on precisely localizing when and which soccer actions related to the ball occur, 
(2) \textit{Dense Video Captioning}, focusing on describing the broadcast with natural language and anchored timestamps, 
(3) \textit{Multi-View Foul Recognition}, focusing on analyzing multiple viewpoints of a potential foul incident to classify whether a foul occurred and assess its severity, 
(4) \textit{Game State Reconstruction}, focusing on reconstructing the game state, \ie, the localization and identification of all players, from broadcast videos onto a 2D top-view map of the field. 
The challenges are illustrated in Figure~\ref{fig:graphical_abstract}.
Compared to previous years~\cite{Giancola2022SoccerNet, Cioppa2024SoccerNetChallenge}, tasks (3) and (4) are novel, introducing more annotations and data, while task (1) was improved with additional classes and annotations. Task (2) remains the same as its previous edition.

\section{Ball Action Spotting}
\label{sec:ball_action_spotting}

\subsection{Task description}

Ball action spotting consists in temporally localizing and identifying ball-related actions across $12$ different classes. In contrast to the previous year that only considered $2$ classes: \textit{pass} and \textit{drive}, labels were extended to $12$ different classes of soccer ball actions: \textit{Pass, Drive, Header, High Pass, Out, Cross, Throw In, Shot, Ball Player Block, Player Successful Tackle, Free Kick, and Goal}. Each action is marked by a single timestamp corresponding to the exact time when an action occurs. In contrast to the SoccerNet Action Spotting Challenge, focusing on global actions, ball-related actions feature a higher density of actions, adding a layer of difficulty.
The data consists of $7$ videos from the English Football League games, each broadcast in 720p quality containing all action timestamp annotations, as well as $2$ extra games for the challenge dataset with segregated annotations to avoid overfitting.

\subsection{Metrics}

Similarly to action spotting, ball action spotting is evaluated using the Mean Average Precision (mAP) metric, which combines both precision and recall, considering some time tolerance $\delta$ (mAP@$\delta$). Particularly, it computes the Average Precision, which is an estimate of the area under the precision-recall curve constructed from different thresholds on the confidence scores of the predictions for each class separately. More precisely, the confusion matrix used to compute the precision and recall point for a particular threshold is constructed as follows: Predictions with a confidence score above the considered threshold and ground-truth timestamps are uniquely matched if they have the same class and if the prediction falls within the tolerance $\delta$ of the ground-truth timestamp. Matched pairs are considered as true positives, unmatched predictions are false positives, and unmatched ground-truth annotations are false negatives. Finally, the Average Precision of each class is averaged into the mAP@$\delta$. For the challenge, the mAP is computed at various time tolerances ($\delta$) and is therefore defined as 
$$mAP@\delta= \frac{1}{|\Gamma|} \sum_{\gamma \in \Gamma}^{} AP(\gamma, \delta)\comma$$
where $\Gamma$ is a set of all possible classes and $AP(\gamma, \delta)$ denotes Average Precision of class $\gamma$ for time tolerance $\delta$. 

The Average-mAP isa nother metric that is used for action spotting and the averages the mAP for different time tolerances using the trapezoid integral estimation.
%
%$$avg-mAP = \frac{1}{|\Theta|} \sum_{\theta \in \Theta}^{} mAP@\theta,$$
%\noindent where $\Theta$  is a set of different time tolerances.
%
Due to the fast nature of the ball events, we evaluate the methods' performance based on the \textit{tight average-mAP} (from $1$ to $5$ seconds) and the mAP at those time tolerances. We chose to rank the methods for the challenge according to the mAP@$1$, which indicates that participants need to localize all ball events very precisely, with one-second tolerance.

\subsection{Leaderboard}

This year, $8$ teams participated in the ball action spotting challenge for a total of $61$ submissions, with an improvement of mAP@$1$ from $56.15$ for the baseline to $73.39$ mAP@$1$ for the winning method. The complete leaderboard may be found in Table~\ref{tab:performance_ball_action_spotting}.

\begin{table}[h]
    \centering
    \caption{
    \textbf{Ball action spotting leaderboard.}
    The main metric for the leaderboard and best performances
    are in bold. Team names with a superscript have
provided a summary that may be found in
Appendix~\ref{app:bas} or in Section~\ref{sub:ballactionspottingwinner} for the winning
team.}
    \begin{tabular}{lcc}
        \toprule
        \bf Participant & \textbf{mAP@1}  & tight avg-mAP \\
        \midrule
        T-DEED$^{BAS-1}$ & \textbf{73.39}  & \textbf{77.25}  \\
        UniBw Munich - VIS$^{BAS-2}$ & 71.35 & 75.24  \\
        FS-TAHAKOM$^{BAS-3}$  &  67.09 &  70.64 \\
        MobiusLabs & 62.53 &  66.13  \\
        AI4Sports$^{BAS-5}$  &  62.44&  65.91 \\
        sota$^{BAS-6}$ &62.44 &   66.19	  \\
        SAIVA$^{BAS-7}$ & 56.74 & 60.58  \\
        Baseline& 56.15 &  60.60 \\
        blueblue111	& 3.66 &  4.05 \\
        \bottomrule
    \end{tabular}
    
    \label{tab:performance_ball_action_spotting}
\end{table}

\subsection{Winner}
\label{sub:ballactionspottingwinner}

The winners for this task are \textit{Artur Xarles, Escalera Sergio, Thomas B. Moeslund, and Albert Clapés}. A summary of their method is given hereafter.

\mysection{BAS-1 - T-DEED: Temporal-Discriminability Enhancer Encoder-Decoder}\\
\textit{Artur Xarles, Sergio Escalera, Thomas B. Moeslund, Albert Clap\'es}
\textit{(arturxe@gmail.com, sescalera@ub.edu, tbm@create.aau.dk, aclapes@ub.edu)}

Our approach adapts the Temporal-Discriminability Enhancer Encoder-Decoder (T-DEED)~\cite{Xarles2024TDEED-arxiv} for the SoccerNet Ball Action Spotting 2024 Challenge. T-DEED uses a 2D backbone with Gate-Shift Fuse (GSF)~\cite{Sudhakaran2023GateShiftFuse} modules to generate per-frame representations with local spatiotemporal information. For longer-term temporal modeling, the subsequent temporally discriminant encoder-decoder refines per-frame tokens, with modules designed to increase their discriminability within the sequence while maintaining high output temporal resolution and using multiple scales. A prediction head, similar to those used in the action spotting literature~\cite{Soares2022Temporally, Xarles2023ASTRA}, produces the final predictions. To cope with limited data, we train T-DEED on both SoccerNet Ball and the original SoccerNet~\cite{Deliege2021SoccerNetv2} datasets simultaneously, using two prediction heads for multitask training. The videos are processed in 100-frame clips at $796\times448$ resolution. The final results come from an ensemble of two models that use different clip sampling strategies: one that samples from all possible clips and the other only from clips that contain actions.

\subsection{Results}\label{sec:ball_action_spotting_result}

Similar to the ball action spotting task from the previous year, this year's challenge continues to present the same difficulties while introducing more complex classification requirements. The task targets rapid, subtle events with minimal visual cues, complicating the differentiation between closely occurring incidents. The dataset available is relatively small compared to the action spotting task, encouraging participants to explore semi-supervised, self-supervised, or transfer learning methods using the $500$ unlabeled broadcast games. A significant enhancement this year is the inclusion of $12$ classes of ball actions, requiring more precise classification. Consequently, simple categorizations like ``pass'' are insufficient as specific pass types must be clearly identified.

The solutions presented by the participating teams include various strategies aimed at enhancing action recognition in soccer videos. Key approaches include the use of gray-scale images to simplify visual processing and emphasize motion detection, integration of 2D and 3D convolutional networks to capture both spatial and temporal features, and the application of transfer learning from related domains to enrich model training. Some solutions leveraged dual-model architectures to amplify data efficiency, while others introduced sophisticated machine learning models like the Temporal-Discriminability Enhancer Encoder-Decoder (T-DEED) that refine discriminability and handle diverse temporal scales for improved precision. These approaches demonstrate a focused attempt to improve action spotting through advanced neural architectures, strategic data handling, and domain-specific enhancements, pointing towards a more effective analysis of sports videos.

\section{Dense Video Captioning}
\label{sec:dense_video_captioning}

\subsection{Task description}

This year marks the second edition of the Dense Video Captioning task, introduced by Mkhallati~\etal~\cite{Mkhallati2023SoccerNetCaption} in 2023. The task focuses on describing long untrimmed football videos with engaging textual comments anchored in time. The first steps therefore consist in localizing specific moments in the video where a caption should be inserted, similar to action spotting, followed by generating natural language sentences to detail the events happening at those times. The SoccerNet-Caption dataset comprises $471$ untrimmed broadcast games at 720p resolution and $25$ fps, with a total of $36{,}894$ anonymized and timestamped captions. Additionally, a challenge set of $42$ extra games with segregated annotations is used for ranking the participants.

\subsection{Metrics}
The metric is the same as the one used in the SoccerNet 2023 dense video captioning challenge. It is adapted from the ActivityNet-captions~\cite{Krishna2017DenseCaptioning} metric for single-anchors. Specifically, the metric considers $30$-second time window around each ground-truth caption. Language similarity is then evaluated using the METEOR metric for each generated caption that falls within this $30$-second tolerance. The evaluation results are averaged across all videos and the entire challenge set.

\subsection{Leaderboard}

This year, $3$ teams participated in the dense video captioning challenge for a total of $18$ submissions, with an improvement of METEOR@$30$ from $19.17$ to $27.08$. The leaderboard may be found in Table~\ref{tab:performance_dense_video_captioning}.

\begin{table}[h]
    \centering
    \caption{\textbf{Dense video captioning leaderboard.} The best performance
    is given in bold. The winning team provided a summary that can be found in Section~\ref{sub:densevideocaptioninggwinner}.}
    \begin{tabular}{lc}
        \toprule
        \bf Participant & \textbf{METEOR@30}  \\
        \midrule
        DeLTA Lab$^{DCV-1}$ & \textbf{27.08}   \\
        zaemon1251-hesty & 21.25    \\
        Baseline& 19.17   \\
        Spring711 & 16.96   \\
        \bottomrule
    \end{tabular}
    
    \label{tab:performance_dense_video_captioning}
\end{table}

\subsection{Winner}
\label{sub:densevideocaptioninggwinner}
\mysection{DVC-1 - End-to-end training of LLMs}\\
\textit{Jamshid Tursunboev, Hyeonmin Choe, Sunghwan Moon, Il-Min Kim, and Jae-Mo Kang}
\textit{(jamshid@knu.ac.kr, iissaacc@knu.ac.kr, sunghwan.moon@knu.ac.kr, ilmin.kim@queensu.ca, jmkang@knu.ac.kr)}

Our video captioning solution adapts the BLIP-2 like framework, utilizing a 4-layer transformer decoder with 512 dimensions. The model processes 30 seconds of pre-extracted visual features as input, along with 8 trainable query tokens. The visual encoder's output tokens are then fed into a pretrained GPT-2 model (base or medium) via a linear projection layer. The entire model, including the LLM, is trained end-to-end during training. We also change the action spotting task from binary to multiclass softmax classification and discard low-confidence spotted actions that are below 0.7 in localization. Our approach achieved a METEOR score of 27.42 and BLEU1 of 44.08 on the test set.
Code: 
% \url{https://github.com/gladuz/soccernet-caption-deltalab} 
\href{https://github.com/gladuz/soccernet-caption-deltalab}
{\uline{https://github.com/gladuz/soccernet-caption-deltalab}}
\textbf{Acknowledgments.} This work was supported in part by the NRF of Korea under Grants 2022R1C1C1003464 and 2022R1A4A1033830, and in part by the IITP of Korea under Grant IITP-2024-2020-0-01808.

\subsection{Results}

Similarly to last year, this task is challenging due to its generative nature. Only three teams participated, but only two of them were able to improve upon the baseline. Particularly, the winning team provided a clever solution that improves both the spotting accuracy by replacing the binary classification with a multi-class softmax classification, as well as improving the text generation quality by fine-tuning end-to-end an LLM. The remaining challenges for this task consist in including the name of the players and the team in the labels used for training and evaluation to provide a true end-to-end solution for automatic commentary generation.

\section{Multi-View Foul Recognition}
\label{sec:multi-view_foul_recognition}

\subsection{Task description}

Multi-View Foul Recognition is a novel computer vision task introduced by Held~\etal~\cite{Held2023VARS} and presented for the first time as a SoccerNet challenge for this year. %that was first presented as a SoccerNet challenges this year. 
The challenge puts the participants in the shoes of a football referee by asking them to classify whether a potential incident is an offence. It requires determining the severity of the foul, defined by four classes: ``No offence'', ``Offence + No card'', ``Offence + Yellow card'', and ``Offence + Red card''. It also requires a classification into one of $8$ fine-grained foul classes (\ie, ``Standing tackling'', ``Tackling'', ``High leg'', ``Pushing'', ``Holding'', ``Elbowing'', ``Challenge'', and ``Dive/Simulation''). 
The task is composed of two classification sub-tasks for the offense and for the severity.
The \textit{SoccerNet-MVFouls} dataset gathers $3{,}901$ actions, each composed of at two, three, or four different synchronized viewpoints of the same scene at 224p and 720p resolution and 25 fps. 

\subsection{Metrics}
As the \textit{SoccerNet-MVFouls} dataset is highly unbalanced, we select the balanced accuracy as the main metric that is defined as follows:
\begin{equation}
    \textrm{\mbox{Balanced Accuracy (BA)}} = \frac{1}{N}\sum_{i=1}^{N} \frac{TP_i}{P_i}\comma
\end{equation}
with $N$ being the number of classes, $TP_i$ (True Positives) is the number of times the model correctly predicted the class $i$ and $P_{i}$ (Positives) is the number of ground-truth samples for that class in the dataset.
The final score for the leaderboard is obtained by taking the mean of the balanced accuracies of the two classification sub-tasks as follows:

\begin{equation}
    combined\_metric = \frac{BA_{type} + BA_{off}}{2}\comma
\end{equation}
with $BA_{type}$ the balanced accuracy for the type of foul classification and $BA_{off}$ for the task of determining if it was a foul and its severity.

\subsection{Leaderboard}
This year, $15$ teams participated in the multi-view foul recognition challenge for a total of $110$ submissions,
with an improvement of the balanced accuracy from $36.99\%$ for the baseline to $44.76\%$ for the winning team. The leaderboard may be found in
Table~\ref{tab:performance_mv_foul}.

\begin{table}[t]
    \centering
    \caption{
    \textbf{Multi-view foul recognition leaderboard}. 
    The main metric for the leaderboard and best performances
    are in bold. Team names with a superscript have
provided a summary that may be found in
Appendix~\ref{app:MVF} or in Section~\ref{sub:multiviewfoulwinner} for the winning
team. Acc. stands for accuracy, BA. stands for balanced accuracy.}
    \resizebox{\linewidth}{!}{% <------ Don't forget this %
    \begin{tabular}{l|lc|lc|c}
         &  \multicolumn{2}{c|}{ Type of Foul} & 
        \multicolumn{2}{c|}{Offence Severity} & \bf Combined metric \\ \midrule
\bf Feat. extr.  & \bf Acc. & \bf BA. & \bf  Acc. & \bf BA. & \bf BA. \\ \midrule
$WJB^{MVF-1}$                &   $48.87$ &   $30.88$ &   $46.15$ &$\bf58.65$ &$\bf44.76$ \\ 
$MLV\_SoccerNet^{MVF-2}$     &   $38.46$ &   $40.94$ &   $36.65$ &   $44.06$ &   $42.50$ \\
$MobiusLab^{MVF-3}$          &   $27.15$ &   $32.46$ &   $38,01$ &   $51.99$ &   $42.23$ \\ 
$UniBw Munich - VIS^{MVF-4}$ &   $52.04$ &$\bf48.70$ &$\bf50.23$ &   $35.40$ &   $42.05$ \\
$hhqweasd$                   &$\bf63.80$ &   $44.58$ &$49.77$ &   $30.48$ &   $37.53$ \\ 
$Baseline$ \cite{Held2024Towards-arxiv} & $1.582$ &   $39.52$ &   $47.51$ &   $34.46$ & $36.99$\\
$pangzihei$                  &   $52.94$ &   $39.98$ &   $45.25$ &   $31.63$ &   $35.81$ \\
$PW ZZSN^{MVF-9}$            &   $42.53$ &   $40.23$ &   $44.34$ &   $26.12$ &   $33.18$ \\
$cmri-ai$                    &   $62.44$ &   $39.05$ &   $41.63$ &   $27.17$ &   $33.11$ \\
$zyz$                        &   $51.58$ &   $35.38$ &   $45.70$ &   $30.41$ &   $32.89$ \\
$t5$                         &   $14.48$ &   $13.55$ &   $39$.37 &   $47.52$ &   $30.53$ \\
$ly001$                      &   $38.91$ &   $33.51$ &   $45.70$ &   $26.71$ &   $30.11$ \\
$Sch$                        &   $16.29$ &   $29.09$ &   $51.58$ &   $29.04$ &   $29.06$ \\
$footballComp$               &   $50.68$ &   $20.95$ &   $44.70$ &   $23.97$ &   $22.46$ \\
$saudi\_zzz$                 &   $12.12$ &   $11.54$ &   $57.92$ &   $25.00$ &   $18.27$ \\\midrule
    \end{tabular}
    }
    \label{tab:performance_mv_foul}
\end{table}

\subsection{Winner}
\label{sub:multiviewfoulwinner}

The winners for this task are Jing Zhang and Xinyu Liu.
A summary of their method is given hereafter.

\mysection{MVF-1 - M2VFCN for multi-view video analysis}\\
\textit{Jing Zhang, Xinyu Liu, Kexin Zhang, Yuting Yang, Licheng Jiao, and Shuyuan Yang\\}
\textit{(\{23171214443,231712144438\}@stu.xidian.edu.cn)}

The team proposed a multi-task classification network named M2VFCN for multi-view video analysis, which significantly enhances classification performance without requiring extensive training time. Initially, the team conducted an in-depth analysis and statistical examination of the dataset, revealing substantial class imbalance issues among different categories in the training set. To address this issue, an adaptive loss function was introduced. Furthermore, to further enhance the model's representation capabilities, the team utilized the Kinetics 400 dataset~\cite{Kay2017TheKinetics-arxiv} for large-scale pretraining. By integrating domain knowledge from general scenarios, the network's representation capacity was significantly augmented.

\subsection{Results}
This novel task introduces several new challenges to the field of action recognition in soccer. One primary difficulty is that visually similar actions may result in entirely different classifications. For example, although two tackles may appear similar, one might be classified as ``No offense'' if there was no contact, while the other could be classified as ``Offense + Yellow card'' if the player made contact with the opponent. 
Small differences, such as the speed of the foul, the point of contact or the intention to play the ball, can result in different classifications.
A second challenge is the subjective nature of the task. The interpretation of what constitutes a foul can vary between referees, leading to inconsistencies. This subjectivity introduces variability and complicates the implementation of an objective classification model.

The participants explored multiple directions for this first edition of the challenge.
First, Hong~\etal used the VideoChat2-based
video foundation model~\cite{Li2024MVBench} as a feature extractor to more effectively capture temporal information.
Rüfenacht adopted a distinct approach by relying only on a single view (the live feed). Given that the portion of the frame covering the foul is small, he employed a lightweight feature extractor to process full-resolution $720\times1280$ grayscale images instead of the commonly used $224\times224$ resolution.

\section{Game State Reconstruction}
\label{sec:game_state_reconstruction}

\subsection{Task description}
SoccerNet Game State Reconstruction (GSR) is a novel computer vision task involving the tracking and identification of players from a single moving camera to construct a video game-like minimap without any specific hardware worn by the players. While this task holds immense value for the sports industry, there was no appropriate open-source benchmark for method comparison until the SoccerNet-GSR dataset was released~\cite{Somers2024SoccerNetGameState}.

Game State Reconstruction can be seen as a compression task, where the goal is to extract high-level information about a sports game given a raw input video. 
The high-level information to be extracted includes the following:
\begin{enumerate}
    \item The 2D position of all sports persons on the field
    \item Their role (player, goalkeeper, referee, or other)
    \item For players and goalkeepers:
    \begin{enumerate}
        \item Their jersey number
    \end{enumerate}
    \begin{enumerate}
        \item Their team affiliation (\ie, left or right w.r.t. the camera viewpoint)
    \end{enumerate}
\end{enumerate}

\subsection{Metrics}
Game State Reconstruction performances are measured by a novel evaluation metric, the GS-HOTA~\cite{Somers2024SoccerNetGameState}, which is an extension of the HOTA metric, a popular evaluation metric for multi-object tracking.
GS-HOTA measures the ability of a GSR method to correctly track and identify all athletes on the sports pitch.
The HOTA metric employs the Intersection-over-Union (IoU) as a similarity score to match predictions (P) and ground truth (G) bounding boxes in the image space.
The key distinction setting GS-HOTA apart from HOTA is the use of a new similarity score, that accounts for the specificities of the GSR task, \ie the additional target attributes (jersey number, role, team) and the detections provided as 2D points in the real-world instead of bounding boxes in the image plane.
This new similarity score, denoted $Sim_{GS-HOTA}(P, G)$, is formulated as follows:

\begin{equation} \label{eq:sim_gs_hota}
Sim_{\text{GS-HOTA}}(P, G) = \text{LocSim}(P, G) \times \text{IdSim}(P, G)\comma
\end{equation}
with
\begin{equation} \label{eq:locsim}
\text{LocSim}(P, G) = e^{\ln(0.05)\frac{\|P - G\|_2^2}{\tau^2}}\comma
\end{equation} 
and
\begin{equation} \label{eq:idsim}
\text{IdSim}(P, G) = 
\begin{cases} 
1 & \text{if all attributes match,} \\
0 & \text{otherwise.}
\end{cases}
\end{equation}

We refer to the work of Somers~\etal~\cite{Somers2024SoccerNetGameState} for more details about the GS-HOTA metric.

\subsection{Leaderboard}

For its first edition, $16$ teams participated in the game state reconstruction challenge with a total of $34$ submissions and an improvement of the GS-HOTA from $23.36$ for the
baseline to $63.81$ for the winning team. The leaderboard
may be found in Table~\ref{tab:performance_gsr}.

\begin{table}[t]
    \centering
    \caption{
    \textbf{Game state reconstruction leaderboard.}
    The main metric for the leaderboard and best performances
    are in bold. Team names with a superscript have
provided a summary that may be found in
Appendix~\ref{app:GSR} or in Section~\ref{sub:gamestatereconstructionwinner} for the winning
team.}
    \resizebox{\linewidth}{!}{% <------ Don't forget this %
    \begin{tabular}{l|c|c|c}
Participant team & \textbf{GS-HOTA (↑)} & GS-DetA (↑) & GS-AssA (↑) \\
    \midrule
    Constructor tech$^{GSR-1}$ & \textbf{63.81} & \textbf{49.52} & \textbf{82.23} \\
    UPCxMobius$^{GSR-2}$ & 43.15 & 30.46 & 61.16 \\
    JAM & 34.40 & 19.38 & 61.08 \\
    XJTU\_MM (JNR) & 33.57 & 20.44 & 55.16 \\
    VIPLab & 29.82 & 16.16 & 55.03 \\
    JustTesting & 29.08 & 13.78 & 61.37 \\
    Robo Space$^{GSR-7}$ & 26.92 & 11.04 & 65.66 \\
    sjc & 24.20 & 10.16 & 57.68 \\
    ABL & 23.73 & 9.94 & 56.67 \\
    testingGS (test baseline) & 23.73 & 9.94 & 56.67 \\
    Baseline~\cite{Somers2024SoccerNetGameState} & 23.36 & 9.80 & 55.69 \\
    SAIVA\_Ball & 23.08 & 10.13 & 52.62 \\
    Football AI Lab & 22.96 & 9.92 & 53.17 \\
    playbox x NUSG$^{GSR-14}$ & 21.26 & 7.55 & 59.90 \\
    UWIPL & 20.33 & 7.85 & 52.64 \\
    eidos.ai$^{GSR-16}$ & 8.92 & 1.51 & 52.69 \\
    Eidos & 8.92 & 1.51 & 52.69 \\
    \bottomrule

    \end{tabular}
    }
    \label{tab:performance_gsr}
\end{table}

\subsection{Winner}
\label{sub:gamestatereconstructionwinner}

The winners for this task are Vladimir Golovkin~\etal.
A summary of their method is given hereafter.

\mysection{GSR-1 - Constructor.tech football multi-object tracking system}\\
\textit{Vladimir Golovkin, Nikolay Nemtsev, Vasyl Shandyba, Oleg Udin, Nikita Kasatkin, Pavel Kononov, Anton Afanasiev, Sergey Ulasen, Andrei Boiarov}
\textit{(\{Vladimir.Golovkin, Nikolay.Nemtsev, c-Vasyl.Shandyba, ou, nk, pak, Anton.Afanasiev, su, Andrei.Boiarov\}@constructor.tech)}

Constructor.tech's football multi-object tracking system identifies athletes and camera parameters for each frame in the input video using the YOLOv5 detector model. The system calculates the ReID embedding, player orientation, TeamID, and Jersey number (if applicable) for all detected athletes. The pitch calibration process involves first obtaining camera parameters from a CNN-Transformer Hybrid model and then refining these parameters using a Field keypoints model (Resnet50, 74 keypoints) through a heuristic voting mechanism based on reprojection error. The tracking pipeline comprises three steps: team detection, raw tracking, and post-processing. The system assigns one of five roles to each player using the TeamID model (OsNet, trained for 111 different uniforms). Tracking is performed using DeepSORT operating in pitch coordinates, with certain restrictions based on player orientation and TeamID. The final stage involves refining and concatenating tracklets into longer tracks, ensuring continuity by matching tracklet start and end times, ReID embeddings, TeamID, Jersey numbers, and player speed.

\subsection{Results}

Despite the first edition of the challenge, participants increased the baseline performance by a large margin.
As discussed in the original SoccerNet-GSR paper \cite{Somers2024SoccerNetGameState}, some modules of the pipeline, including the YOLOv8 object detector for player detection and the MMOCR module for digit recognition, were not fine-tuned on soccer data and were therefore a bottleneck to the overall performance. As a result, most participants employed fine-tuned object detectors (YOLOv8, YOLOv5m, \etc) and jersey number recognition models to increase the overall performance. Additionally, some participants implemented the role classification layer as part of the object detector.
Various approaches were introduced for calibration, such as a SegFormer-based one-step calibration method and camera calibration technique introduced by the ``No Bells, Just Whistles'' paper \cite{GutierrezPerez2024NoBells}.
Improvements in jersey number detection included the usage of a YOLOv8-based detector with 0-99 classes or a digit detector with 0-9 classes combined into numbers. Some teams opted for jersey number recognition without detection using a WideResNet architecture.
%\textbf{Continue with other bullet points:}

% Athlete detector fine-tuning on soccer data, YOLOv8 & YOLOv5m
% Role classification (player, gk, referee) by the object detector
% SegFormer one-step calibration
% Camera calibration from “No Bells, Just Whistles”
% Number detector with 0-99 classes (YOLOv8)
% Digit detector with 0-9 classes combined into numbers
% JN recognition without detection (WideResnet)
% ReID : many different models (BPBReID, PRTReID, Person-ReID)
% Player orientation (New module)
% Accelerated using DeepStream & TensorRT

\section{Conclusion}
\label{sec:conclusion}

This paper summarizes the outcome of the SoccerNet 2024 challenges. In total, we present the
results on four tasks: ball action, dense video captioning, multi-view foul recognition, and game state reconstruction. These challenges provide a comprehensive overview of current state of-the-art methods within each computer vision
task. For each challenge, participants were able to significantly improve the performance over our proposed baselines or previously published state of the art. Some tasks such as ball action spotting are reaching promising
results for industrial use, while novel tasks such
as multi-view foul recognition and game state reconstruction may still require further investigation. In the future, we will keep on extending the set of tasks, challenges, and benchmarks related to video understanding in sports.

\textbf{Acknowledgement.} A. Cioppa is funded by the F.R.S.-FNRS. 
This work was partly supported by the King Abdullah University of Science and Technology (KAUST) Office of Sponsored Research through the Visual Computing Center (VCC) funding and the SDAIA-KAUST Center of Excellence in Data Science and Artificial Intelligence (SDAIA-KAUST AI).

\clearpage

%%%%%%%%% REFERENCES
{\small
%\bibliographystyle{ieee_fullname}
%\bibliography{bib/abbreviation-short,bib/action,bib/camera-calibration-sports,bib/dataset,bib/labo,bib/learning,bib/soccer,bib/soccernet-challenge, bib/sports,bib/NEW-REFS-HERE}}

\clearpage

\section{Supplementary Material}
\label{sec:supplementary_material}

\subsection{Ball Action Spotting}
\label{app:bas}

\mysection{BAS-2 - Transformer for long range dependencies in action recognition}

\textit{Konrad Habel, Fabian Deuser, Norbert Oswald}
\textit{(\{konrad.habel, fabian.deuser, norbert.oswald\}@unibw.de)}

Our solution employs a two-stage approach, with stage-1: a visual features extractor inspired from the baseline solution and stage-2: a Transformer for long-range interrelationships between actions. For the feature extractor, we use as backbone an EfficientNetV2 B0 and stack 21 gray-scaled frames at 720p resolution. Using every second frame at 25 fps results in a 1.68 seconds context window for stage-1. For stage-2 we apply a Transformer with sequence length 768 and use every second extracted feature of stage-1, what leads to a context window of 61.44 seconds for stage-2. An ensemble of three different models trained with different games as validation holdout (valid, test-1, test-2) achieves on the test set a mAP@1 score of 70.13 and on the challenge set a mAP@1 score of 71.35.

\mysection{BAS-3 - FS-TAHAKOM}\\
\textit{Faisal Sami Altawijri and Saad Ghazai Alotaibi}\\
\textit{(faltawijri@tahakom.com, sgalotaibi@tahakom.com)}

Two models with identical structures were employed, initially trained with transfer learning from the ``Action Spotting Challenge'' due to limited data (7 videos). The first model, initialized with weights from a pretrained model on 500 videos, achieved 64.5 mAP. Further enhancements included optimizing frame inputs (27 frames) and focal loss parameters. The second model, also pretrained on football videos, contributed domain-specific knowledge, aiding in feature extraction. A multi-stage training process followed: individual training of each model and then fine-tuning their concatenated architecture, freezing most layers to prevent overfitting. This strategy integrated datasets from both challenges, improving feature extraction and achieving the final score of 67.0 mAP.

\mysection{BAS-5 - AI4Sports}\\
\textit{Linyi Li and Haobo Wang\\}
\textit{(\{2153274, 2153067\}@tongji.edu.cn)}

Our solution is mainly based on the official baseline of this challenge. This model architecture uses a slow fusion approach. It uses 2D convolution block to extract 2D features at early stages, and 3D convolution block to extract 3D features at later stages. 
Our improvements focus on adjusting the hyperparameters of the baseline architecture. 
First, we tried adjusting the initial sampling weights. The label of other 10 classes is far fewer than pass and drive, so the 10 classes should have a higher possibility to be sampled. So we tried various ways to adjust the initial sampling weights: 
One minus class frequency: \[ w_i = 1-\frac{N_i}{N} \point \] 
Reciprocal of class frequency: \[ w_i = \frac{1}{N_i} \point \] 
Reciprocal of the square root of class frequency: \[ w_i = \frac{1}{\sqrt{N_i}} \comma \] 
and the $w_i$ will be multiplied by 12. The reciprocal of the square root of class frequency worked best. 

\mysection{BAS-6 - sota Team}\\
\textit{Zhihao	Li, Pei	Geng, Yuxuan	Xiao, Jian	Yang, and Shanshan	Zhang}
\textit{(zhihaoli0828@gmail.com,
 peigeng99@gmail.com,
 xiaoyuxuan@njust.edu.cn,
 csjyang@njust.edu.cn,
 shanshan.zhang@njust.edu.cn)}

Our model follows the 1st place solution for the SoccerNet Ball Action Spotting Challenge 2023. The model is based on a slow fusion approach that synergistically integrates 2D and 3D convolutions to capture spatial and temporal dynamics within video frames. The model consumes sequences of grayscale frames, allowing the model to focus on the structural elements of the frames.  Considering the long-tail distribution problem, we use Focal Loss to alleviate the problem of unbalanced action classes and achieve optimal performance with $\alpha=0.4$ and $\gamma=1.2$.  During training, we employ 7-fold cross-validation to ensure robust generalization across the dataset. During prediction, the horizontal flipping is used to increase the model confidence.

\mysection{BAS-7 - SAIVA}\\
\textit{WonTaek Chung, HanKyul Kim, and ByoungKwon Lim}\\
\textit{(wwchung91@gmail.com, harry.kim@aibrain.com, bklim@aibrain.com)}

Our approach extends the 2D and 3D convolution architecture presented by Baikulov~\cite{Cioppa2023SoccerNetChallenge-arxiv} with a transformer. The proposed architecture extracts spatial features from 2D convolutions and temporal features from 3D convolutions, creating spatial temporal patches. However, the spatial temporal patches only represent the sequential images locally because 2D convolutions have small receptive fields to describe the topological data and 3D convolutions can only represent short intervals of the sequential data. The proposed architecture uses a transformer to create a global representation of an image sequence. The 2D convolution layers and 3D convolution layers use pretrained weights from the action spotting task and compute the spatial temporal patches. The transformer encoder layers take the spatial temporal patches as input and create a global representation of an image sequence. A linear classifier uses the output of the transformer encoder layers to determine the ball action of the sequential image data.

%\subsection{Dense Video Captioning}
%\label{app:DVC}

\subsection{Multi-View Foul Recognition}\label{app:MVF}

\mysection{MVF-2 - VF-VARS}\label{MVF-2}\\
\textit{Jihwan Hong, Youngseo Kim, Junseok Lee, Jimin Lee, Sehyung Kim, and Hyunwoo J. Kim}
\textit{(csjihwanh@korea.ac.kr, xwsa568@korea.ac.kr, behindstarter42@korea.ac.kr, 2001joe@korea.ac.kr, shkim129@korea.ac.kr, hyunwoojkim@korea.ac.kr)}

Our VF-VARS method integrates the VideoChat2-based video foundation model~\cite{Li2024MVBench} and VARS~\cite{Held2023VARS} to improve the performance of temporal pooling-based vision encoders. To handle a variable number of video clips, we ensured each video has four clips. The first clip shows an overall field view, while the others provide close-up shots from the SoccerNet-MVfoul dataset~\cite{Held2023VARS}. When only two clips are available, we duplicate the second clip twice with different augmentations. When three clips are given, we randomly select and duplicate either the second or third view with augmentations. We tackled the suboptimal performance by using the VideoChat-2 model, which is known for effectively capturing temporal information. For classification, videos are fed into the pre-trained UMT-L~\cite{Li2023Unmasked} video encoder, which we freeze to prevent forgetting. Visual features are then converted into query tokens using QFormer~\cite{Li2023BLIP2}. We fine-tuned QFormer with definitions of fouls and specific instructions. Final classifications are obtained through a 3-layered MLP. Our code is available at \href{https://github.com/csjihwanh/soccernet-MLV}{https://github.com/csjihwanh/soccernet-MLV}.

\mysection{MVF-3 - Single-View Foul Recognition}\label{MVF-3}\\
\textit{Dominic Rüfenacht}\\
\textit{(dominic.ruefenacht@gmail.com)}

Perhaps the most interesting aspect of the proposed method for foul recognition is that it solely relies upon view 0 of the multi-view dataset (\ie, the live feed). As the portion of the frame in view 0 that covers the foul is small, we opted for a lightweight feature extraction backbone that can consume full-resolution $720\times1280$ grayscale images as input. These are stacked into “fake” RGB images, before being passed through a sequence of 2D and 3D convnets that captures spatio-temporal relationships~\cite{Baikulov2023Solution}. The output of the backbone is a 1280 dimensional vector, which is passed through a number of MLPs. First, there is a shared MLP for both offense severity and action, the output of which is then passed through individual MLPs for the two classification tasks. Horizontal flipping is used as test-time augmentation, which improved results by around 1.5\% points. The full code is available on \url{https://github.com/druefena/MVFoul}~\cite{ RuefenachtVARS2024}.

\mysection{MVF4 - TAdaFormer for foul recognition}

\textit{Konrad Habel, Fabian Deuser, Norbert Oswald}
\textit{(\{konrad.habel, fabian.deuser, norbert.oswald\}@unibw.de)}

Our approach leverages a TAdaFormer-L/14~\cite{Huang2023TemporallyAdaptive-arxiv} pre-trained on the Kinetics 400 and 710 datasets~\cite{Kay2017TheKinetics-arxiv}. We use an ensemble of three models using frame 75 as center frame with 3 different step sizes (1, 2, 3), what leads to a context window of (16, 32, 48) frames. The model is kept frozen until Transformer stage 21. We only train the last three Transformer stages and the used classification heads on the foul recognition data. Similar to the baseline solution~\cite{Held2023VARS}, we use max-pooling on the output of the TAdaFormer to aggregate multiple views before the classification heads. During training, we randomly select two views, whereas during inference all views per action are used. An ensemble of three models trained with different step size between consecutive frames achieves on the test set a combined metric score of $42{.}37$ and on the challenge set a combined metric score of $42{.}05$.

\mysection{MVF9 - PW ZZSN}

\textit{Milosz Lopatto and Adam Gorski}
\textit{(\{milosz.lopatto,adamsebastiangorski\}@gmail.com)}

We based our solution on the existing VARS model with the MViTv2 encoder~\cite{Li2022MViTv2}. First, we refactored and wrapped the code in Pytorch Lightning, which allowed us to easily track experiments using a custom Weights and Biases logger. We experimented with multiple parameters, but FPS, start frame, and end frame are the ones that have been changed compared to the baseline model configuration. We kept the middle of the extracted clip at the 75th frame, but we changed the FPS to 12 instead of 17 proposed in the baseline model. We have also experimented with data augmentation. Besides making existing transformations more aggressive, we have also added RandomResizedCrop and GaussianBlur.
These changes led us to a 3rd place on the public test leaderboard. With more time, we would try the Hiera architecture and compare it to the MViTv2. 
\textbf{Acknowledgments.} Our research was supported in part by PL-Grid Infrastructure grant nr PLG/2024/017176.

\subsection{Game State Reconstruction}
\label{app:GSR}

\mysection{GSR-2 - No bells, More Whistles}\\
\textit{Marc Gutiérrez-P\'erez and Dominic R\"ufenacht}\\
\textit{(dominic.ruefenacht@gmail.com)}

Our submission to the SoccerNet gamestate reconstruction challenge uses the Tracklab library \cite{Joos2024Tracklab} with three significant modifications to improve performance. First, we replaced the camera calibration method with the one from Gutierrez \etal~\cite{GutierrezPerez2024NoBells}, which involves a hierarchical keypoint generation process leveraging the soccer field's geometric properties, and uses an HRNetv2-based encoder-decoder network for keypoint and line detection. Second, we retrained the person detection model on around 2000 custom-annotated frames from SoccerNet videos and directly classified the detections as players, goalkeepers, and referees. Lastly, we trained a dedicated jersey digit detection model on publicly available data and around $1{,}000$ custom annotated frames from SoccerNet videos. The recognized digits are subsequently merged to form the jersey numbers. This submission significantly improves over the baseline, addressing core issues effectively and enhancing the overall methodology.

\mysection{GSR-7 - Robo Space}\\
\textit{Zehua Cheng, Yuan Li, Wei Dai, Menglong Li, and Yongqiang Zhu}
\textit{(zehua.cheng@cs.ox.ac.uk, lyn@intellicloud.ai, loveispdvd@gmail.com, mlli8803@163.com, alexzhu.vip@gmail.com)}

The methodology outlined commences with utilizing TrackLab as the core framework due to its proven effectiveness in analyzing behaviors and tracking, establishing a solid foundation. To improve object detection accuracy specific to the research needs, YOLOv8 is fine-tuned using a custom dataset, focusing on precise localization of humans and their actions in videos. In conjunction, the Body Pose-Based Re-Identification (BPBReID) model is refined to strengthen individual recognition even in visually complex situations, contributing to consistent tracking. To enhance the system's understanding of dynamic scenes over time, a Global Temporal Attention mechanism is introduced, which selectively highlights important spatiotemporal elements, improving the interpretation of behaviors. Finally, time smoothing techniques are applied to refine detections, eliminating noise and ensuring coherence throughout the video, facilitating seamless analysis of continuous actions and paths. This comprehensive strategy combines established and innovative methods, from the solid base of TrackLab to the specialized tuning of YOLOv8 and BPBReID, integration of temporal attention, and concluding with time smoothing.

\mysection{GSR-14 - playbox x NUSG}\\
\textit{Atom Scott and Calvin Yeung}\\
\textit{(atom.james.scott@gmail.com, yeung.chikwong@g.sp.m.is.nagoya-u.ac.jp)}

We fine-tuned YOLOv8X for detection while retaining the original tracking model implemented in tracklab. Camera calibration was improved by correcting a tvcalib implementation error which excluded the center circle when calculating segment reprojection loss and applying temporal filtering, using roll degrees as an error proxy since broadcast cameras should not exhibit roll. For jersey number recognition, we used a two-stage CLIP-based method: one model detected numbers in bounding boxes, another identified specific numbers. This approach, fine-tuned on a custom dataset (we manually annotated $3{,}000$ bounding boxes with jersey numbers) achieved 0.78 tracklet accuracy. GPT-4O was used to refine predicted roles, team affiliation and jersey numbers. Despite qualitative improvements, especially in camera calibration, and a 4\% increase in test set scores over the baseline, our methods didn’t surpass the baseline on the final challenge set. We expect camera calibration and jersey number recognition to have significant room for improvement.

\mysection{GSR-16 - Further than aspect: solving roles with expanded information context}\\
\textit{Francisco Cach\'on, Enzo Pacilio, Cristian Gonzalez, and Marcelo Ortega}\\
\textit{(\{francisco.cachon, enzo, cristian, ortegatron\}@eidos.ai)}

Our approach is focused on improving role and team classification, solving it globally for the video instead of classifying each track individually. We trained a Siamese Network to produce bounding box embeddings, and cluster these embeddings together with different KMeans values. To find the most plausible clustering and assign role and team to its clusters, we trained a Random Forest classifier, on features incorporating additional game context information derived from detected tracks and its fieldmap position. This method achieved a lower GS-HOTA, likely because the model was trained with ground truth positional information, but finally evaluated on inaccurate detection data.

\end{document}